\documentclass[letterpaper, 10 pt, conference]{ieeeconf}  

\IEEEoverridecommandlockouts                              

\usepackage{cite}
\usepackage{amsmath,amssymb,amsfonts}
\usepackage{algorithmic}
\usepackage{graphicx}
\usepackage{textcomp}
\usepackage{xcolor}
\usepackage{enumerate}
\usepackage[linesnumbered,ruled,lined]{algorithm2e}
\usepackage{booktabs}
\usepackage{threeparttable}
\usepackage{multirow}
\usepackage{stfloats}
\usepackage{subfigure}
\usepackage{graphicx}
\usepackage{rotating}

\title{\LARGE \bf
Accurate and Robust Object-oriented SLAM with 3D Quadric Landmark Construction in Outdoor Environment
}

\author{Rui Tian$^{1}$, Yunzhou Zhang$^{1*}$, Yonghui Feng$^{1}$, Linghao Yang$^{1}$, \\Zhenzhong Cao$^{1}$, Sonya Coleman$^{2}$, Dermot Kerr$^{2}$
	\thanks{$^*$The corresponding author of this paper. }
	\thanks{$^{1}$Rui Tian, Yunzhou Zhang, Yonghui Feng, Linghao Yang and Zhenzhong Cao are with College of Information Science and Engineering, Northeastern University, Shenyang 110819, China (Email: {\tt\small zhangyunzhou@mail.neu.edu.cn}).}%
	\thanks{$^{2}$Sonya Coleman and Dermot Kerr are with School of Computing, Engineering and Intelligent Systems, Ulster University, N. Ireland, UK.}
	\thanks{This work was supported by National Natural Science Foundation of China (No. 61973066, 61471110)  and the Distinguished Creative Talent Program of Liaoning Colleges and Universities (LR2019027).}
}

\begin{document}

\maketitle
\thispagestyle{empty}
\pagestyle{empty}

\begin{abstract}
Object-oriented SLAM is a popular technology in autonomous driving and robotics. In this paper, we propose a stereo visual SLAM with a robust quadric landmark representation method. The system consists of four components, including deep learning detection, object-oriented data association, dual quadric landmark initialization and object-based pose optimization. State-of-the-art quadric-based SLAM algorithms always face observation related problems and are sensitive to observation noise, which limits their application in outdoor scenes. To solve this problem, we propose a quadric initialization method based on the decoupling of the quadric parameters method, which improves the robustness to observation noise. The sufficient object data association algorithm and object-oriented optimization with multiple cues enables a highly accurate object pose estimation that is robust to local observations. Experimental results show that the proposed system is more robust to observation noise and significantly outperforms current state-of-the-art methods in outdoor environments. In addition, the proposed system demonstrates real-time performance.
\end{abstract}

\section{INTRODUCTION}

Simultaneous Localization and Mapping (SLAM) is a fundamental technique in order for robots to perceive the environment. When compared with classic SLAM methods that use only the geometry of the scene, object-based SLAM has recently focused on creating maps with both geometry and high-level semantic objects within the environment\cite{rubino20173d,ok2019robust,2018Monocular,yang2019cubeslam,wu2020eao,nicholson2018quadricslam,2020Object,V2020Perspective,hosseinzadeh2019real}. This semantically-enriched information can help robots with target-oriented tasks like obstacle avoidance, robust relocalization and human-robot interaction. The improvement in the accuracy of semantic information acquisition, driven by deep learning networks\cite{redmon2018yolov3,bolya2019yolact,he2017mask}, has led to the increasing introduction of object detection and semantic segmentation into visual SLAM systems to build semantically enriched maps and enhance the perception ability of robots.

\begin{figure}[ht]
\centering
 	\includegraphics[scale=0.42]{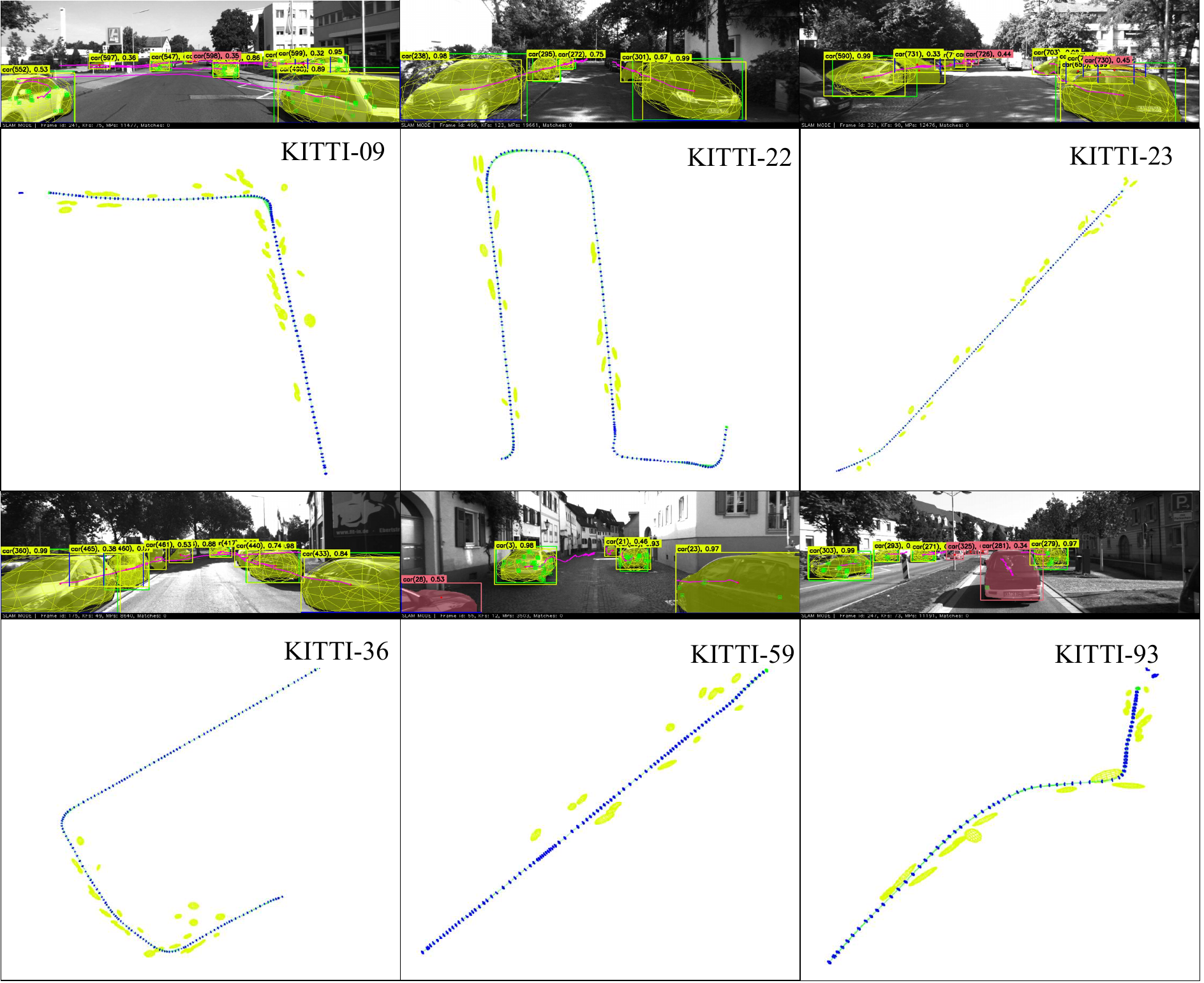}
 	\caption{Our proposed method uses 3D quadric landmarks to build the object-oriented map in outdoor environments. Yellow quadrics illustrate the accuracy of orientation and shape of the estimated ellipsoids when projected onto the current image frame.  Object ID is marked, and the magenta lines show the center of the ellipsoids in previous frames when projected onto the image frame, indicating the accuracy of the object data association. The red bounding box is the object detected as dynamic and will not be 
 constructed with ellipsoids. Finally, the object maps are also provided. }
 	\label{kitti}
 	\vspace{-5mm}
 \end{figure}

Accurate object representation is a key issue in object-oriented SLAM research and 3D object models\cite{salas2013slam++}, cubic boxes\cite{wu2020eao,yang2019cubeslam,2018Monocular} and ellipsoids\cite{nicholson2018quadricslam,2020Object,V2020Perspective} are among common methods utilized for object representation. Prior work like \cite{yang2019cubeslam} and \cite{wu2020eao} use the cubic box to represent the object, where the pose of the cubic box can be estimated by vanishing points and rotation sampling. Compared with the cubic box, the ellipsoid can also accurately represent the position, orientation and size of the object and has a more concise mathematical representation\cite{nicholson2018quadricslam}. In projective geometry the quadric can be represented by a symmetric matrix\cite{cross1998quadric} where the compact perspective projection model and the closed surfaces of ellipsoids are meaningful for object landmarks.

The accuracy and robustness of current quadric-based SLAM are not ideal, especially the quadric initialization process, which is limited by the parameter coupling of the direct linear solution method\cite{nicholson2018quadricslam} or the necessity for point cloud fitting\cite{2020Object,hosseinzadeh2019real}.  QuadricSLAM\cite{nicholson2018quadricslam} is a recently proposed object-oriented SLAM system that represents objects as quadrics; a dual quadric observation model based on the object detection is proposed. 
However, the closed-form constrained dual quadric parameterization and the lack of observation angles under the planar trajectory of the mobile vehicle make the initialization of the quadric difficult and sensitive to observation noise. In \cite{hosseinzadeh2019real}, multiple constraints combined with points, surface and quadrics are used in the optimization framework, but the prior shape of the object is estimated based on deep learning which incurs a high computational complexity and is not robust. In \cite{ok2019robust}, the texture plane and shape prior constraints are added to the quadric estimation which solves the problem of poor estimation performance when the observation angles change in road driving scenes. However, the assumption that the texture plane is parallel to the image plane during quadric initialization causes the estimation to be sensitive to noise.

In addition, in prior work such as \cite{nicholson2018quadricslam,2020Semantic}, data association methods have been proposed although they are typically not robust to outdoor scenes. Dynamic objects in outdoor scenes like moving cars and persons are a challenge for quadric estimation since false object associations will lead to false quadric initialization results.

To solve the aforementioned problems, we propose a robust and accurate quadric landmark initialization based on a method for decoupling of quadric parameters (DQP) and an object data association (ODA) algorithm in outdoor scenes. The robustness of DQP to observation noise is improved by independently estimating the quadric centroid translation and the yaw rotation constraint which is satisfied for autonomous vehicles in road planes in most cases. Then, an ellipsoid with improved accuracy can be obtained by a nonlinear optimizer combining the observation error, the texture plane error and the prior object size. In terms of data association, we propose a multiple-cues algorithm combined with the Hungarian assignment algorithm\cite{kuhn1955hungarian} which improves the robustness of object pose estimation.

We demonstrate the performance of the proposed system in both a simulation environment and using the KITTI Raw Data \cite{Geiger2013IJRR} datasets. The experimental results show that the proposed system is more robust to observation noise than other existing methods and improves the accuracy of the position, orientation and size of the object estimation in the outdoor environment.

\textbf{The main contributions of this work are:}
\begin{itemize}
	\item  To effectively overcome the observation noise, we propose an accurate and robust quadric landmark initialization method based on the DQP algorithm by decoupling of translation and rotation of quadric centroids.   
	\item We proposed an ODA algorithm that combines the semantic inliers distribution, Kalman-based motion prediction, and ellipsoidal projection to achieve accurate object data association and object pose estimation.
	\item Based on the proposed algorithms, we implement real-time stereo visual SLAM with accurate and robust ellipsoids representing objects, aiming to build an object-oriented and semantically-enhanced map for outdoor navigation.
\end{itemize}

\section{System Overview}

\subsection{Mathematical Representation of a Quadric Model}
For convenience of description, the notations used in this paper are as follows:
\begin{itemize}
\item $(\cdot)_w$ is the world coordinate, $(\cdot)_c$ is the camera coordinate, $(\cdot)_r$ is the reference camera coordinate of the object,  and $(\cdot)_q$ is the quadric center frame.

\item $K$ - The intrinsic matrix of a pinhole camera model.

\item $T_{cw}\in{R^{4\times4}}$ - The transformation from world frame to camera frame, which is composed of a rotation $R_{cw}\in{R^{3\times3}}$ and a translation $t_{cw}\in{R^{3\times1}}$.

\item $P=KT$ - The camera projection matrix that
contains intrinsic and extrinsic camera parameters.

\item $B=[x_1, y_1, x_2, y_2]^T$ - The 2D object detection bounding box (BBox).


\item $M$ is the segmentation instance mask, $D$ is the detection instance, $O$ is the object instance.

\item $D^i_j$ represent the detected object instance $D_j$ that is assigned to the object $O_i$, $cls(D)$ and $cls(O)$ represent the class label of the detected instance and object instance respectively. 

\item $In(B, x)$ - The checking of image points $x$ that are located in the $B$ detection box.

\item $Q\in R_{4\times4}$ - The quadric matrix in 3D space and $Q^*\in R_{4\times4}$ is denoted as the dual quadric matrix. 

\item $\Pi=[\pi_1, \pi_2, \pi_3, \pi_4]^T$ - The 3-D plane surface in homogeneous coordinate and all quadric plane fulfil $\Pi^TQ^*\Pi=0$.

\item $q=[a_x, a_y, a_z, t_x, t_y, t_z, \theta_x, \theta_y, \theta_z]^T$ - The 9-D vector representing the attributes of the quadric, including axial length, translation and rotation. 
\end{itemize}

When a dual quadric is projected onto an image plane, it creates a dual conic, following the rule $C^*=PQ^*P^T$. For more specific properties of the quadric, please refer to \cite{nicholson2018quadricslam}.

\begin{figure}[]
\centering
 	\includegraphics[scale=0.43]{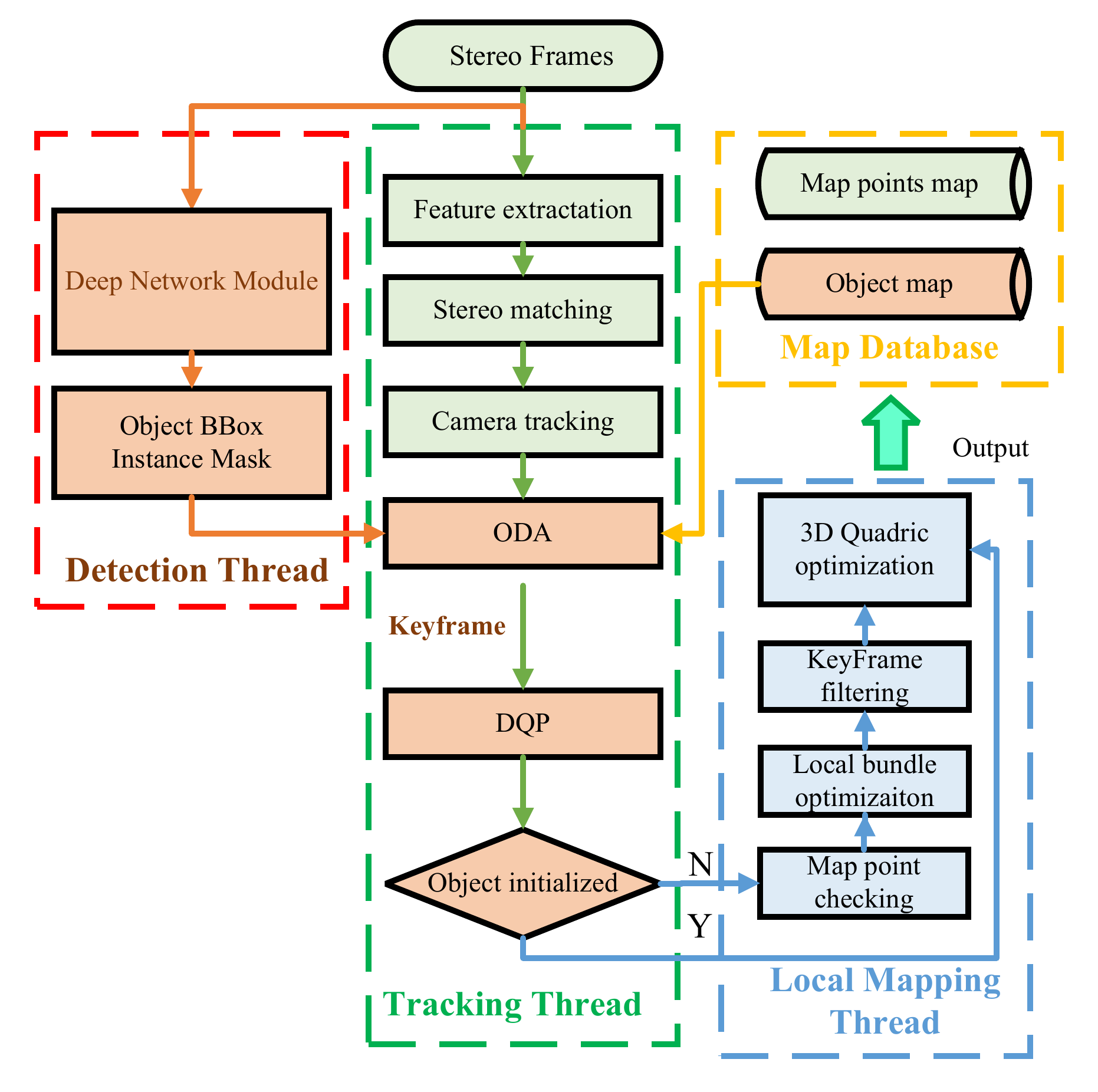}
 	\caption{Overview of our proposed system, there are two key modules: 
 	1) The detection thread takes images and acquires semantic and detection results. 
 	2) The tracking thread initializes quadrics with the DQP method and the ODA associates detected object quadrics with the mapped objects for further optimization. Finally, the object map is stored with ellipsoid representations. }
 	\label{overview}
 	\vspace{-3mm}
 \end{figure}
 
\subsection{System Architecture}
The proposed system is shown in Fig.\ref{overview}. We implement our algorithms on the basis of ORB-SLAM3 \cite{2020ORB}, and a stereo camera is used to obtain a metric scale of the estimated trajectory for the autonomous driving scene to avoid scale ambiguity caused by monocular SLAM\cite{2021Accurate}.  However, we also highlight that our method can be used for monocular SLAM. There are two key modules, the visual SLAM module and the detection module. The visual SLAM module consists of parallel threads, including the tracking thread and the local mapping thread. Finally, the camera pose is estimated and a semantically-enhanced object map is also stored in the map database.

(1) \textbf{The detection thread} uses YOLOACT\cite{redmon2018yolov3} to acquire semantic information from the left images of the stereo pair. The output results are object detection BBoxes and the instance segmentation masks.

(2) \textbf{The tracking thread} takes images and estimates the camera pose from consecutive frames. Meanwhile, the thread waits for the detection instances and associates them with the existing objects in the object map database or decides whether to create a new object using the ODA algorithm. In addition, if the current frame is a keyframe and an observation satisfies the quadric initialization condition, the DQP algorithm is used for robust and accurate quadric initialization.

(3) \textbf{The local mapping thread} optimizes the map points of keyframes with local bundle adjustment. In addition, when the objects are observed by newly inserted keyframes, the new observation can be added to the object optimizer for nonlinear optimization of the ellipsoidal representation of objects.

(4)	\textbf{The map Database} stores the final maps, including the geometry information of map points and the object-oriented map with ellipsoids.

\section{Decoupling of Quadric Parameters Initialization Algorithm} 

\subsection{Decoupling of Quadric Central Translation}\label{AA}

We present the mathematical analysis of the dual quadric parameters to illustrate the effect of the translation component on the estimation of rotation and shape.  The dual form parameters of the ellipsoid can be decomposed by eigen-decomposition in the reference camera coordinates of the object: 
\begin{equation}
\begin{aligned}
    Q_r^*&=T_{rq}Q_{q}^*T_{rq}^T\\
&    =
    \begin{bmatrix}
    R_{rq} & t_{rq} \\
    0^{T} & 1
    \end{bmatrix}
    \begin{bmatrix}
    D & 0 \\
    0^{T} & -1
    \end{bmatrix}
    \begin{bmatrix}
    R_{rq}^T & 0 \\
    t_{rq}^{T} & 1
    \end{bmatrix}\\
&    = \begin{bmatrix}
    R_{rq}DR_{rq}^T-t_{rq}t_{rq}^T & -t_{rq} \\
    -t_{rq}^{T} & -1
    \end{bmatrix}\\
&    =\begin{bmatrix}
    Q^*_{r33} & -t_{rq} \\
    -t_{rq}^{T} & -1
    \end{bmatrix}
    \end{aligned}
    \label{eq1}
\end{equation}
\\
where $D\in{R^{3\times3}}$ is the diagonal matrix composed of the squares of the quadric axial lengths, and $t_{rq}\in{R^{3\times1}}$ is the quadric centroid translation in the reference camera coordinates. The parameters of the block matrix $Q^*_{r33}\in{R^{3\times3}}$ couple the rotation and translation of the quadric. Since the length of the quadric centroid translation is much larger than that of the rotation and axes, small errors in the estimation of the quadric centroid translation have a significant impact on the accurate estimation of the dual quadric matrix, which is why QuadricSLAM \cite{nicholson2018quadricslam} is sensitive to observation noise.

We can also see from Eq.\ref{eq1} that the translation parameters are independent in dual form parameters $q_{14}^*$, $q_{24}^*$, $q_{34}^*$. Therefore we estimate the translation component parameters $t_{rq}\in{R^{3\times1}}$ independently to eliminate the effect of coupling parameters, a key aspect of our approach.  We triangulate the center of the 2D detection box $x_{bc}^i$ and obtain the triangulation map point $\hat{t}_{rq}$, which is almost close with the quadric center $t_{rq}$ in outdoor scenes. This assumption is proved by experiments in \ref{QSIM}. Observations of two or more frames of detection centers form the overdetermined equation to solve $t_{rq}$, 
\begin{equation}
\left[\begin{array}{l}
P_{c}\left[1,{ }^{*}\right] T_{w r}-x_{bc}^i[1] \cdot P_{c}[3,{ }^{*}] T_{w r} \\
P_{c}\left[2,{ }^{*}\right] T_{w r}-x_{bc}^i[2] \cdot P_{c}[3,{ }^{*}] T_{w r}
\end{array}\right] t_{rq}=0
\end{equation}

where, $x_{bc}^i[i]$ is the $i$-th element of 2D detection center, $P_c[i,*]$ is the $i$-th row of the projection matrix $P_c$.

\subsection{Decoupling of Quadric Rotation and Axial Length}
The rotation and quadric axial lengths are considered after the quadric centroid translation has been estimated independently. We assume that the ellipsoid of the object, such as an autonomous vehicle or robot, is under the constraint of yaw rotation, while the pitch and roll are constant at zero. This is satisfied for autonomous vehicles on the road in outdoor scenes. Therefore, we can replace the rotation matrix in Eq.\ref{eq1} by:

\begin{equation}
\begin{aligned}
        R_{rq}&=\begin{bmatrix}
        cos\theta_y & 0 & sin\theta_y\\
        0 & 1 & 0\\
        -sin\theta_y & 0 & cos\theta_y
        \end{bmatrix}\\
        Q_{r}^*&=
        \resizebox{.8\hsize}{!}{$
        \begin{bmatrix}
        a_x^2cos^2\theta_y+a_z^2sin^2\theta_y-{t_{rq}^{x2}} & -t_{rq}^xt_{rq}^y & cos{\theta}_{y}sin{\theta}_{y}(a_z^2-a_x^2)-t_{rq}^xt_{rq}^z & -t_{rq}^x\\
        -t_{rq}^xt_{rq}^y & a_y^2-{t_{rq}^{y2}} & -t_{rq}^yt_{rq}^z & -t_{rq}^y \\ cos{\theta}_{y}sin{\theta}_{y}(a_z^2-a_x^2)-t_{rq}^xt_{rq}^z & -t_{rq}^xt_{rq}^y & a_{x}^2sin^2\theta_y+a_z^2cos^2\theta_{y}-{t_{rq}^{z2}} & -t_{rq}^z\\
        -t_{rq}^x & -t_{rq}^y & -t_{rq}^z & -1
        \end{bmatrix}
        $}
        \end{aligned}
\end{equation}
		
where, $t_{rq}^x$, $t_{rq}^y$, $t_{rq}^z$ are elements of the quadric centroid translation vector.

 We can simplify the linear form in \cite{nicholson2018quadricslam} by using the landmark BBox observations $B$ and the corresponding dual quadric planes $\Pi$ by substituting the $t_{rq}$.
\begin{small}
\begin{equation}
\begin{aligned}
&M^{T}=
\begin{bmatrix}
\pi_1^2\\
2\pi_1\pi_3\\
\pi_2^2\\
\pi_3^2\\
\pi_4^2+2\pi_1\pi_2t_{rq}^xt_{rq}^y+2\pi_1\pi_4t_{rq}^x+
2\pi_2\pi_3t_{rq}^yt_{rq}^z\\+2\pi_2\pi_4t_{rq}^y+2\pi_3\pi_4t_{rq}^z
\end{bmatrix}
\end{aligned}
\end{equation}
\end{small}

\begin{equation}\label{eqM}
    M
    \begin{bmatrix}
    q_{11}^* & q_{13}^* & q_{22}^* & q_{33}^* & q_{44}^*
    \end{bmatrix}^T=0
\end{equation}

The decoupled linear form of Eq.\ref{eqM} can be solved by singular value decomposition (SVD) \cite{nicholson2018quadricslam}, where $q_{ij}$ is the remaining elements of the dual quadric to be estimated.

Finally, the 9-D vector $q$ of the quadric with orientation, translation and axial lengths of the ellipsoid can be obtained by the estimated dual quadric matrix $Q^*$:

\begin{equation}
\begin{aligned}
&\theta_y=arctan(2Q_3/(Q_8-Q_1))/2\\
&a_x=\sqrt{\left|Q_1-2Q_3+Q_8\right|/2}\\
&a_y=\sqrt{Q_2}\\
&a_z=\sqrt{\left|Q_1+2Q_3+Q_8\right|/2}
\end{aligned}
\end{equation}

where,
\begin{equation}
\begin{aligned}
Q_1&=-\frac{q^*_{11}}{q^*_{44}}+{t^{x2}_{rq}} \quad Q_2=-\frac{q^*_{22}}{q^*_{44}}+{t^{y2}_{rq}}\\ Q_3&=-\frac{q^*_{13}}{q^*_{44}}+t^x_{rq}t^z_{rq} \quad Q_8=-\frac{q^*_{23}}{q^*_{44}}+{t^{z2}_{rq}}\\
\end{aligned}
\end{equation}


\section{3D Object Observation Constraints Optimization}
In the local mapping thread, we optimize the quadrics by using odometry factors and landmark factors combined with the observation of local keyframes. We define the set of detected objects as $\mathbb{D}$, and the set of mapped objects as $\mathbb{O}$. By minimizing the observation error between observed instances $D_k$ and associated mapped instance $O_i$, $q$ of the quadric can be optimized with the following constraint:

\begin{equation}
\begin{aligned}
&q=\mathop{\arg\min}_{q}(\sum H_b(f_b)+\sum H_a(f_a))+\sum H_p(f_p)\\
    \end{aligned}
\end{equation}

The Huber kernel $H(.)$ is used to enhance the robustness of outlier observations, and the $LM$ algorithm is used to optimize the target cost function. 


\subsection{The 2D detection error} 
The 2D detection error is used to calculate the distance error between the 2D object BBox $B^k_{O_i}$ and the detected BBox $B_{D_{j}}$ in the $k^{th}$ keyframe.
Detection results near the edge of the image are ignored in order to eliminate the effect of occlusion.
\begin{equation}
    \begin{aligned}
    &f_b=e_b(B^k_{O_i},B_{D_{j}})^T\Omega_be_b(B^k_{O_i},B^{D_{j}})\\
    &e_b(B^k_{O_i},B_{D_{j}})=B^k_{O_i}-B_{D_{j}}
    \end{aligned}
\end{equation}
		
\subsection{Prior axial length error} 
The prior axial length error is calculated by the distance between the prior axial length $a_{prior}$ and the object quadric axial length $a_{O_i}$ with the same object class.
\begin{equation}
 \begin{aligned}
f_a&=e_a(O_i)^T\Omega_ae_a(O_i)\\
e_a(O_i)&=a_{prior}(cls(O_i))-a_{O_i}
\end{aligned}
\end{equation}

\subsection{Texture plane error} 
Similar to the method proposed by \cite{ok2019robust}, the texture plane error is obtained by the minimum distance between the fitted texture plane and the quadric landmark. The plane parameters of the texture plane is obtained by Delauney Triangulation of the object's map points with the normal vector $n_{D_j}$ and plane distance $Z_{D_j}$ of a texture plane $\Pi_{D_j}$. The texture plane distance error can be calculated as:
\begin{equation}
\begin{aligned}
& f_p=e_p(\Pi_{D_j},\Pi_{O_i})^T\Omega_pe_p(\Pi_{D_j},\Pi_{O_i})\\
& e_p(\Pi_{Dj},\Pi_{O_i})=Z_{D_{j}}-Z_{O_i}
\end{aligned}
\end{equation}

\section{ The Object Data Association algorithm}
Multi-view geometry information is used for object landmark initialization, while the object detection results are obtained by the single-frame image. Therefore, it is necessary to correctly associate the detected instance of the same object within the map. We propose the ODA algorithm to integrate information for data association. The Hungarian algorithm \cite{kuhn1955hungarian} is used to complete the assignment with the minimum distance error. Three different distance metrics are used for affinity functions to obtain $a_{ij}$, which is the element of the cost matrix $\mathbb{A}$. The $\alpha$, $\beta$ and $\gamma$ parameters are experimentally set to 0.8, 1, and 0.8 respectively.   
\begin{equation}
    \begin{aligned}
    a_{ij}=\alpha{a^p_{ij}}+\beta{a^g_{ij}}+\gamma{a^k_{ij}}
    \end{aligned}
\end{equation}

\subsection{Semantic Inliers Points Distance}
To overcome the overlap of the detection masks, we use Bi-directional Optical Flow (BODF) to track the keypoints within the detection mask $M_j$ from the last keyframe and obtain the keypoints set $\{x^k_j\}$. We calculate the ratio of inliers corresponding to the same object class, where $\text{size}\{.\}$ calculates the element numbers of the set:

\begin{equation}
    a^{p}_{ij}=1-\frac{\text{size}\{ In (M^{k}_{O_i},X_j^{k})\}}{\text{size}\{X_j^{k}\}}
\end{equation}

\subsection{Intersection of Union Distance}
To calculate the intersection of union distance, we use the intersection ratio between the 2D quadric landmark projection BBox $B_{Oi}$ of $O_i$ and the 2D detection result $B_{D_j}$ of the object instance $D_j$ .

\begin{equation}
    a^g_{ij}=1-\frac{B_{O_i}\cap{B_{D_j}}}{B_{O_i}\cup{B_{D_j}}}
\end{equation}

\subsection{Prior Object Size Distance}
For each object instance $O_i$, the motion prediction method based on the Kalman filter \cite{2016Simple} is adopted to predict the state of the detection in the image frame. The predicted 2D BBox of $D_j$ is denoted as $B^a_{D_j}$, the prior object size distance $a^k_{ij}$ is defined by:
\begin{equation}
    a^k_{ij}=1-\frac{B^a_{O_i}\cap{B^a_{D_j}}}{B^a_{O_i}\cup{B^a_{D_j}}}
\end{equation}


\section{Experiments}
The proposed system consists of two modules, including the SLAM module and the detection module. The overall system architecture is described in Fig.\ref{overview}. In order to evaluate the performance of our proposed method, we build an experimental simulation environment based on OpenGL to compare the robustness and accuracy against other state-of-the-art techniques. The KITTI Raw Data dataset\cite{Geiger2013IJRR} is adopted as the benchmark real-world dataset to demonstrate the effectiveness of our method in outdoor scenes. All the experiments are conducted using an Intel(R) Core(TM) i7-9750H CPU@2.6GHZ, 16G memory, and Nvidia GTX 1080 Ti.

\textbf{We define the following criteria for evaluation:}

\textbf{(1)} $IoU_{2D}$: The intersection ratio between the ground truth (GT) and the estimated quadric projection detection.
\begin{equation}
    IoU_{2D}=\frac{B_{gt}\cap{B_{pred}}}{B_{gt}\cup{B_{pred}}}
\end{equation}

\textbf{(2)} $e_{trans}$: The error of quadric centroid translation between the GT ellipsoid and the prediction estimation, indicating the accuracy of the ellipsoid position estimation.
\begin{equation}
    e_{trans}=\|t^{gt}_{wq}-t^{pred}_{wq}\|_2
\end{equation}

\textbf{(3)} $e_{axe}$: The error of ellipsoid axial length between the GT ellipsoid and the predict estimation in the world coordinate, indicating the accuracy of the object shape estimation.
\begin{equation}
    e_{axe}=\|a_{gt}-a_{pred}\|_2
\end{equation}

\subsection{Quantitative Evaluation of Simulation}\label{QSIM}
Simulation provides GT of object positions and it is easy to test the robustness of the methods with different types of disturbance. We create the synthetic dataset with OpenGL, five cameras are evenly deployed within $18^{\circ}$ circular arcs to simulate the camera observation in the outdoor environment. An ellipsoid with varying shape and yaw rotation is deployed, the GT 2D object BBox and the position are provided. 
The yaw rotation of the ellipsoid is randomly sampled in the range of $\pm 5^{\circ}$ to simulate objects with rotation. To avoid the influence of random errors on the experimental results, for each type of noise, 10 ellipsoids are generated with Gaussian noise from 10 seeds resulting in a total of 100 trials.

To test the effect of different types of noise on the quadric initialization method, the relative camera poses are obtained by introducing zero-mean Gaussian noise with standard deviations in the range $5\%\sim 30\%$ to simulate the trajectory error. In addition, a detection BBox is simulated by adding the zero-mean Gaussian noise of $1\%\sim 6\%$ to the GT.

We compare methods of quadric initialization including (a) Nicholson \textit{et al} \cite{nicholson2018quadricslam} denoted as Q-SLAM, (b)  Rubino \textit{et al} \cite{rubino20173d} denoted as Conic-method, (c) the proposed method with only decoupling of the quadric central translation, denoted as Tri, and (d) the proposed initialization method denoted as Tri+Yaw.

\begin{figure*}[]
\centering
 	\includegraphics[scale=0.30]{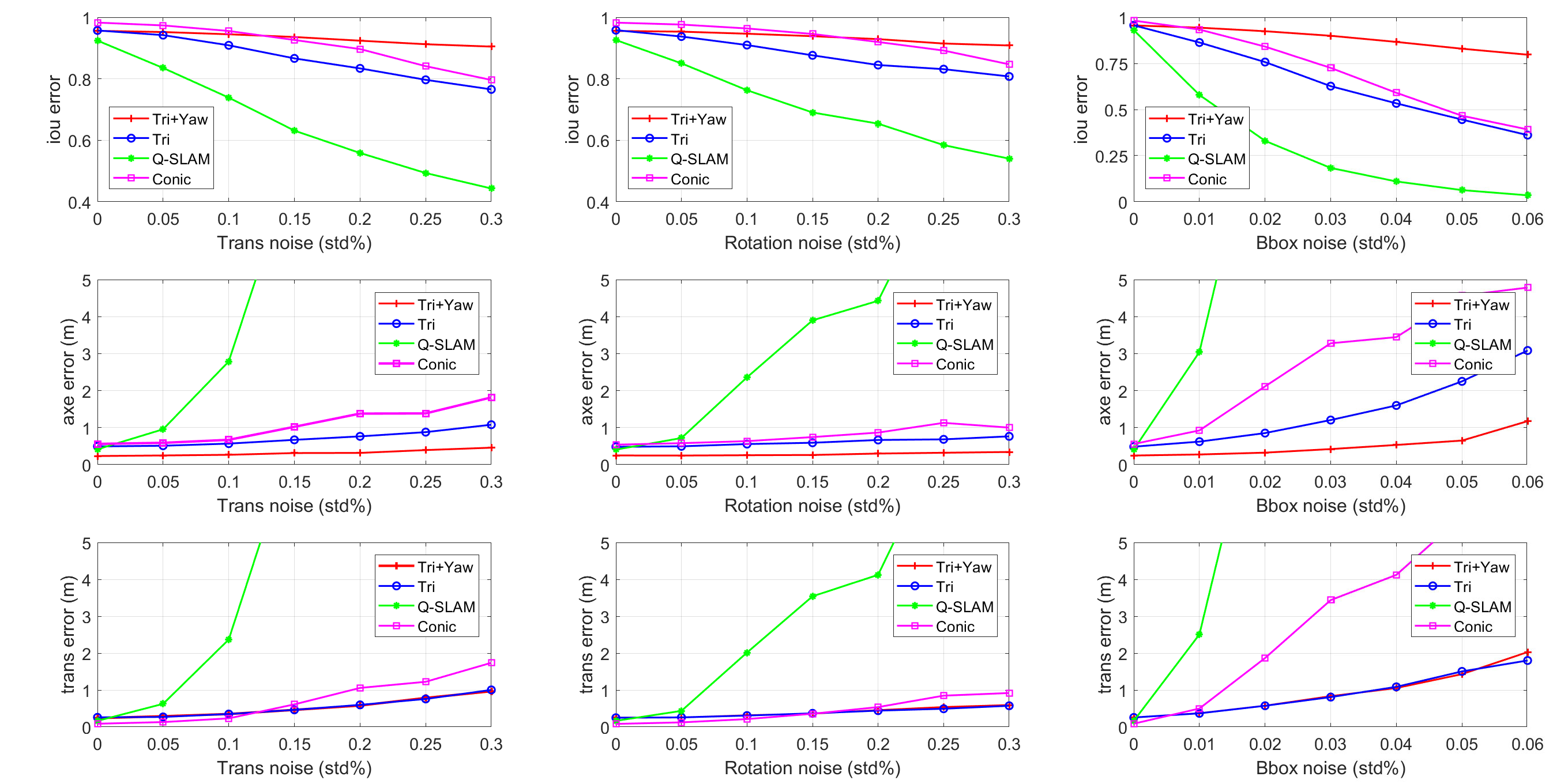}
 	\caption{The initialization performance of methods to different types of noise, the curves show the trend of different criteria with the increase of noise.}
 	\label{com}
 	\vspace{-5mm}
 \end{figure*}

Quantitative evaluation results of initialization methods with different types of noise are visualized in Fig.\ref{com}. The plots show the trend of different evaluation criteria with the increase in noise. It can be seen from Fig.\ref{com}, the results for all methods are consistent with the GT, demonstrating correctness of all methods with zero noise. The performance of all methods degrades when noise increases.

It is obvious that the Q-SLAM method is the most sensitive to noise among all the techniques. When either the translation noise reaches 15\%, the rotation noise reaches 20\%, or the detection BBox noise reaches 2\%, Q-SLAM fails to construct ellipsoids. 

\begin{figure}[]
\centering
 	\includegraphics[scale=0.42]{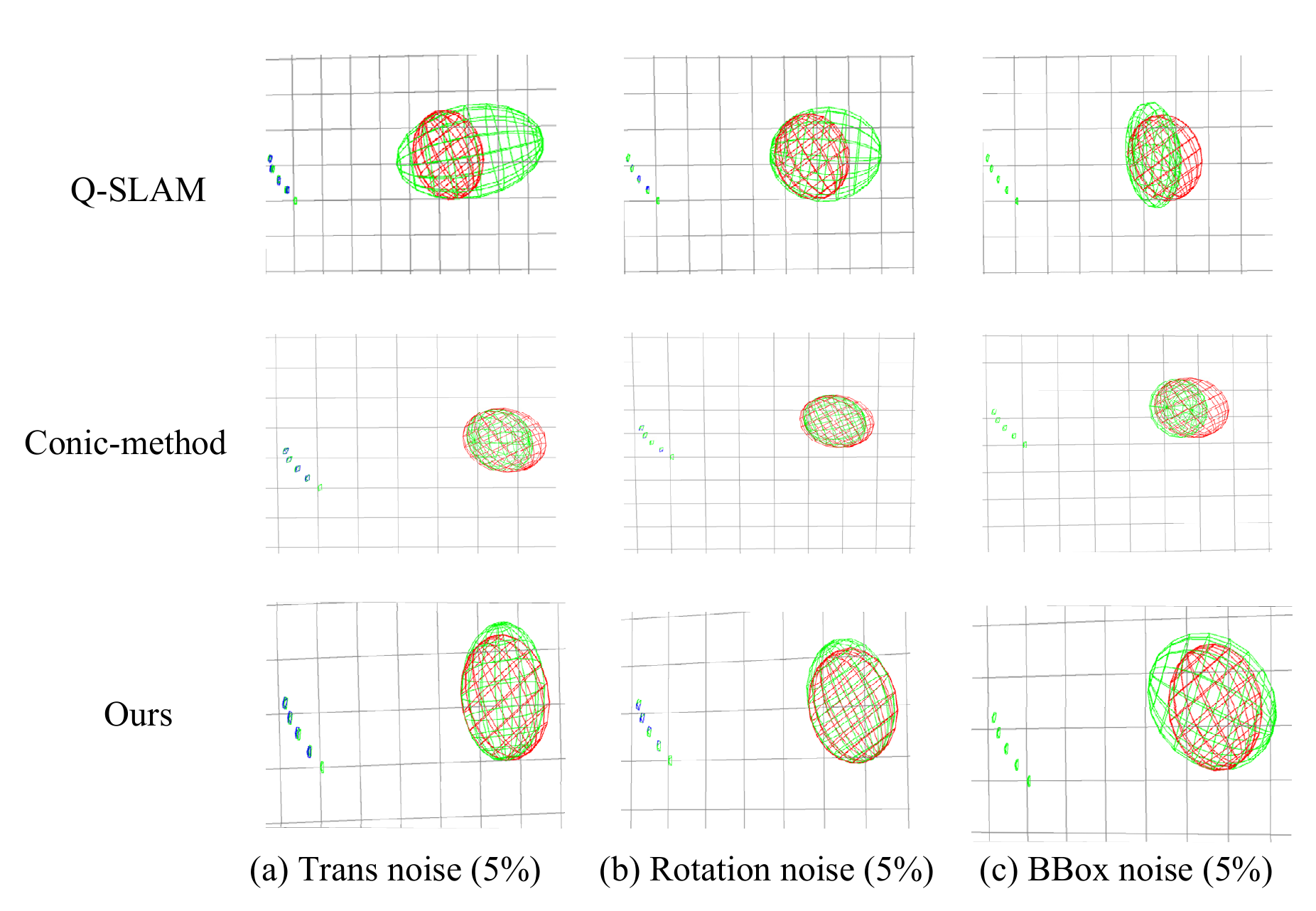}
 	\caption{The visualization results of initialization performance of methods to different types of noise. Our method can initialize ellipsoid accurately and robustly.}
 	\label{simvisul}
 \end{figure}

Meanwhile, the Conic-method maintains relatively good results which show the robustness under the effect of translation and rotation noise. On the other hand, it can be seen that under the influence of detection BBox noise, the $e_{trans}$ and $e_{axe}$ of the Conic-method also increase rapidly. When the detection BBox error exceeds 4\%, the Conic-method fails to initialize the ellipsoids, which indicates that the Conic-method is also sensitive to detection noise. However, the performance of the proposed method is stable as it can be seen that with translation and rotation noise, the error remains stable with the maximum axial error of 0.45m and maximum translation error of 0.89m. These metrics are also influenced within a small range by the detection BBox error with the maximum axial error and translation error of 1.02m and 2.10m. These results show that our proposed method significantly improves the robustness of initialization with the minimal growth trend of noise. The visualization results of quadric initialization are shown in Fig.\ref{simvisul}, where the red ellipsoid is the GT, and green ellipsoid is the estimation. Our proposed method outperforms all compared methods.

\begin{table}[h]
\begin{center}
\caption{Success rate comparison in KITTI Raw Data dataset.}\label{table_success}
\setlength{\belowcaptionskip}{-0.5cm}   
\setlength{\tabcolsep}{5mm}{
\begin{tabular}{c|c c c}
\hline
Sequence & Ours & Conic\cite{rubino20173d} & Q-SLAM\cite{nicholson2018quadricslam}\\
\hline
09 & \textbf{0.6912} & 0.4468 & 0.2706\\
22 & \textbf{0.6512} & 0.3333 & 0.2923\\
23 & \textbf{0.6230} & 0.3949 & 0.2829\\
36 & \textbf{0.6047} & 0.4096 & 0.3514\\
59 & \textbf{0.5625} & 0.3556 & 0.2558\\
93 & \textbf{0.4815} & 0.2826 & 0.3617\\
\hline
Average & \textbf{0.6023} & 0.3705 & 0.3024\\
\hline
\end{tabular}
}
\end{center}
\vspace{-5mm}
\end{table}

\begin{table}[h]
\begin{center}
\caption{$IoU_{2D}$ comparison in KITTI Raw Data Dataset.}\label{table_iou}
\setlength{\belowcaptionskip}{-0.5cm}   
\setlength{\tabcolsep}{4mm}{
\begin{tabular}{c|c c c}
\hline
Sequence & Ours & Conic\cite{rubino20173d} & Q-SLAM\cite{nicholson2018quadricslam}\\
\hline
09 & \textbf{0.7335} & 0.7252 & 0.7031\\
22 & 0.7629 & \textbf{0.7791} & 0.7662\\
23 & 0.7509 & \textbf{0.7529} & 0.6959\\
36 & \textbf{0.7604} & 0.7558 & 0.7127\\
59 & 0.6508 & \textbf{0.6878} & 0.6500\\
93 & 0.7232 & 0.6751 & \textbf{0.7433}\\
\hline
Average & \textbf{0.7303} & 0.7293 & 0.7119\\
\hline
\end{tabular}
}
\end{center}
\vspace{-5mm}
\end{table}

\subsection{Evaluation on KITTI Raw Data Dataset}
To evaluate the performance of the proposed method in outdoor environments, we select the KITTI Raw Data dataset \cite{Geiger2013IJRR} in particular the sequences -09, -22, -23, -36, -59, and -93, which were recorded in urban and residential areas with vehicles. The dataset provides GT for vehicles, including 3-DoF object size and 6-DoF object pose. With the extrinsic parameters of sensors, we can transform the object pose to camera coordinates.

Table \ref{table_success} shows the success rate of initialization and ellipsoid construction by different methods using different sequences. Tables \ref{table_iou}, \ref{table_trans} and \ref{table_axe}  show the experimental results of successfully constructed ellipsoids under different evaluation criteria.

From Table \ref{table_success}, we can see that our method constructs ellipsoids for 60.2\% of the vehicles and reaches an increase of 62.6\% (from 37.0\% to 60.23\%) and 99.2\% (from 30.24\% to 60.23\%) in success rate compared with the Conic-method and the Q-SLAM, respectively, thus confirming the effectiveness of our initialization method. For the $IoU_{2D}$ metric, larger values indicate better construction results. As can be seen from Table \ref{table_iou}, our method outperforms the other existing methods with respect to $IoU_{2D}$ in sequence -09 and -36, with the overall best average of 73.03\%. The compared methods give better results for individual sequences because they discard some detection results that fail to be initialized. For $e_{trans}$ and $e_{axe}$, smaller values indicate better construction results. As can be seen from Table \ref{table_trans} and Table \ref{table_axe}, our method outperforms the compared methods in all cases except for sequence-22, with the average ellipsoid central translation error of 2.127m, nearly 52.2\% reduction in error. In addition, our average axial length error is 0.642 m, a 50.8\% reduction in error, compared with 1.369 and 0.947 for the other techniques. These experimental results show the robustness and effectiveness of the proposed method for ellipsoid representations in outdoor scenes.

Finally, we show the constructed object maps in Fig.\ref{kitti}. The yellow ellipsoids in the map represent static vehicles and the yellow quadrics illustrate the orientation and shape of the estimated ellipsoids when projected onto the image frame. The  magenta lines show the center of the ellipsoids in previous frames projected onto the current image frame, demonstrating the accuracy of the ODA algorithm. The red BBox represents the vehicles that are detected as dynamic objects and are not contained in the map.

\begin{table}[]
\begin{center}
\caption{Translation error comparison in KITTI Raw Data Dataset.}\label{table_trans}
\setlength{\belowcaptionskip}{-0.5cm}   
\setlength{\tabcolsep}{4mm}{
\begin{tabular}{c|c c c}
\hline
Sequence & Ours & Conic\cite{rubino20173d} & Q-SLAM\cite{nicholson2018quadricslam}\\
\hline
09 & \textbf{2.5456} & 2.7819 & 3.6110\\
22 & 2.1769 & \textbf{1.9552} & 1.9651\\
23 & \textbf{2.3341} & 5.6605 & 8.7088\\
36 & \textbf{1.8594} & 2.7175 & 6.9814\\
59 & \textbf{1.3276} & 1.6883 & 1.3874\\
93 & \textbf{2.5226} & 5.4130 & 4.0654\\
\hline
Average & \textbf{2.1277} & 3.3694 & 4.4532\\
\hline
\end{tabular}
}
\end{center}
\vspace{-5mm}
\end{table}

\section{Conclusion}
In this work, a novel pipeline of real-time object-oriented stereo visual SLAM with 3D quadric landmarks is presented. A quadric initialization method based on the DQP algorithm is proposed to improve the robustness and success rate of ellipsoid construction. The data association is solved by the ODA algorithm which ensures highly accurate object pose estimation. Extensive experiments are conducted to show that the proposed system is accurate and robust to observation noise and significantly outperforms other methods in an outdoor environment.

In further work, we will explore finding the semantic relationships between object ellipsoids, and using the semantic information of the object map to localize and perform re-localization.

\begin{table}[]
\begin{center}
\caption{Axial length error comparison in KITTI Raw Data Dataset.}\label{table_axe}
\setlength{\belowcaptionskip}{-0.5cm}   
\setlength{\tabcolsep}{4mm}{
\begin{tabular}{c|c c c}
\hline
Sequence & Ours & Conic\cite{rubino20173d} & Q-SLAM\cite{nicholson2018quadricslam}\\
\hline
09 & \textbf{0.6271} & 1.2618 & 1.1400\\
22 & \textbf{0.5565} & 0.7233 & 0.8356\\
23 & \textbf{0.5494} & 1.7886 & 0.6837\\
36 & \textbf{0.7121} & 1.2797 & 0.8799\\
59 & \textbf{0.6706} & 1.8467 & 1.2908\\
93 & \textbf{0.7357} & 1.3156 & 0.8574\\
\hline
Average & \textbf{0.6419} & 1.3693 & 0.9479\\
\hline
\end{tabular}
}
\end{center}
\vspace{-5mm}
\end{table}

\enlargethispage{-13.5cm}

\bibliographystyle{IEEEtran}
\bibliography{ref}



\end{document}